On the front panel you will find several 2mm banana sockets with different
colors. Their functions are briefly explained below.\\
   1. 5V OUT - This is a regulated 5V power supply that can be used for
      powering external circuits. It can deliver only upto 100mA current ,
      which is derived from the 9V unregulated DC supply from the adapter.\\
   2. Digital outputs - four RED sockets at the lower left corner . The
      socket marked D0* is buffered with a transistor; it can be used to drive
      5V relay coils. The logic HIGH output on D0 will be about 4.57V
      whereas on D1, D2, D3 it will be about 5.0V. D0 should not be used in
      applications involving precise timing of less than a few milli seconds.\\
 3. Digital inputs - four GREEN sockets at the lower left corner. It might
    sometimes be necessary to connect analog outputs swinging between
    -5V to +5V to the digital inputs. In this case, you MUST use a 1K
    resistor in series between your analog output and the digital input pin.\\
 4. ADC inputs - four GREEN sockets marked CH0 to CH3\\
 5. PWG - Programmable Waveform Generator\\
 6. DAC - 8 bit Digital to Analog Converter output\\
 7. CMP - Analog Comparator negative input, the positive input is tied
    to the internal 1.23 V reference.\\
 8. CNTR - Digital frequency counter (only for 0 to 5V pulses)\\
 9. 1 mA CCS - Constant Current Source, BLUE Socket, mainly for Re-
    sistance Temperature Detectors, RTD.\\
10. Two variable gain inverting amplifiers, GREEN sockets marked IN and
    BLUE sockets marked OUT with YELLOW sockets in between to insert
    resistors. The amplifiers are built using TL084 Op-Amps and have a
    certain offset which has to be measured by grounding the input and
    accounted for when making precise measurements.\\
11. One variable gain non-inverting amplifier. This is located on the bot-
    tom right corner of the front panel. The gain can be programmed by
    connecting appropriate resistors from the Yellow socket to ground.\\
12. Two offset amplifiers to convert -5V to +5V signals to 0 to 5V signals.
    This is required since our ADC can only take 0 to 5V input range.
    For digitizing signals swinging between -5V to +5V we need to convert
    them first to 0 to 5V range. Input is GREEN and output is BLUE.\\